\documentclass{article}

\usepackage{arxiv}
\renewcommand{\headeright}{}
\renewcommand{\undertitle}{}
\usepackage[utf8]{inputenc}
\usepackage[T1]{fontenc}
\usepackage{hyperref}
\usepackage{url}
\usepackage{booktabs}
\usepackage{amsfonts}
\usepackage{amsmath}
\usepackage{amssymb}
\usepackage{nicefrac}
\usepackage{microtype}
\usepackage{graphicx}
\graphicspath{{./images/}}
\usepackage{xcolor}
\usepackage{subcaption}
\usepackage{cleveref}
\crefname{algocf}{Algorithm}{Algorithms}
\Crefname{algocf}{Algorithm}{Algorithms}
\usepackage{natbib}
\usepackage{tikz}
\usetikzlibrary{shapes.geometric, arrows.meta, positioning, fit, backgrounds, calc, shadows.blur}

\usepackage[ruled,vlined,linesnumbered]{algorithm2e}

\usepackage{listings}
\hypersetup{
  pdftitle  = {SwarmHarness: A Decentralized Incentivized Protocol for
               Skill-First Agent Compute Networks},
  pdfauthor = {Edwin Jose},
  pdfsubject= {Decentralized AI, Swarm Intelligence, Agent Harness,
               Incentive Mechanisms, MCP, Federated Compute},
  colorlinks = true,
  linkcolor  = blue!60!black,
  citecolor  = teal,
  urlcolor   = blue!70!black,
}

\title{SwarmHarness: Skill-Based Task Routing via Decentralized Incentive-Aligned AI Agent Networks}

\author{
  Edwin Jose \\
  Department of Computer Science \\
  Western Michigan University \\
  Kalamazoo, MI 49008, USA \\
  \texttt{edwin.jose@wmich.edu}
}

\date{May 2026}

\begin{document}

\maketitle

\begin{abstract}
Vast quantities of compute (GPU cycles on personal workstations, idle
inference servers, and edge devices between jobs) go unused because no
incentive-aligned protocol exists for their owners to share them safely
and profitably.
Existing approaches either require a trusted central coordinator (cloud
marketplaces), demand heavy blockchain infrastructure (Golem,
BrokerChain), or lack an incentive layer entirely (BOINC, Petals).
We propose \textbf{SwarmHarness}, a decentralised protocol in which
HarnessAPI skill nodes self-organise into a compute swarm without any
central authority.
SwarmHarness has three interlocking components: a \emph{SwarmRegistry}
built on a Distributed Hash Table (DHT) for peer discovery and capability
advertisement; a \emph{SwarmRouter} that dispatches tasks to nodes using
a utility function over capability, load, latency, and trust; and
\emph{SwarmCredit}, an incentive mechanism that attributes compute-credit
rewards to contributing nodes via a Shapley-value approximation.
Nodes earn credits by serving tasks and spend credits to submit them;
idle nodes that never contribute drain credits and lose routing priority,
creating a self-regulating participation economy.
As nodes specialise toward high-reward skills and routing signals act as
digital pheromones, the network exhibits emergent collective intelligence
analogous to biological swarms.
Beyond compute sharing, SwarmHarness is a foundational primitive for
autonomous distributed AI agent networks in which agents hire compute,
route subtasks, and settle credits without human intermediation.
\end{abstract}

\section{Introduction}
\label{sec:intro}

Trillions of GPU-hours go unused every year.
A researcher's workstation is idle overnight; an edge server waits between
inference requests; a student's gaming GPU sits dormant between training runs.
Meanwhile, organisations queue expensive cloud compute for tasks that a
nearby, willing machine could execute in seconds.
The barrier is not technical, as CPUs and GPUs can accept remote workloads
today, but \emph{incentive}: owners have no safe, automated mechanism to
expose idle compute, track their contribution, and receive proportional reward.

We previously proposed HarnessAPI \citep{Jose2026harnessapi}, a skill-first
framework that exposes a typed \emph{skill folder} simultaneously as an
HTTP endpoint and a Model Context Protocol (MCP) tool, eliminating the
dual-maintenance burden for LLM tool deployment.
HarnessAPI nodes are already being deployed on personal machines and servers.
Each node is an autonomous agent: it knows its own skills, accepts tasks, and
streams results.
The natural next question is: \emph{what happens when thousands of
HarnessAPI nodes run on hardware owned by different people?}
Beyond mere compute sharing, those nodes could form the substrate
for autonomous AI agents that hire compute, delegate subtasks, and settle
payments without any human or cloud intermediary.

Today, no protocol answers either question without either a central coordinator
or blockchain overhead.
Cloud marketplaces require trust in a platform.
Volunteer-computing systems like BOINC \citep{Anderson2019boinc} harvest idle
cycles but offer no monetary or credit incentive.
Petals \citep{Borzunov2022petals} enables collaborative inference over commodity
GPUs but has no incentive layer, allowing free-riders to consume without contributing.
Federated Learning (FL) frameworks \citep{McMahan2016fedavg} distribute model
training but remain data-centric, not skill-centric, and similarly lack
general-purpose incentive attribution.
Blockchain-based markets (Golem \citep{Golem2018whitepaper},
BrokerChain \citep{Lin2025brokerchain}) provide incentives but impose
on-chain transaction costs and latency incompatible with interactive inference.

We propose \textbf{SwarmHarness}, a decentralised protocol that fills this
gap by layering three components on top of existing HarnessAPI nodes:
a DHT-based \emph{SwarmRegistry} for peer discovery, a utility-scoring
\emph{SwarmRouter} for task dispatch, and \emph{SwarmCredit}, a
Shapley-value incentive mechanism that rewards contribution fairly without
a blockchain.
The resulting network is a \emph{swarm}: no central controller, locally
rational agents, and emergent collective behaviour arising from the
accumulation of credit and trust signals, mirroring the stigmergic dynamics
found in biological swarms.

\paragraph{Contributions.}
\begin{itemize}
  \item We define the SwarmHarness system model, formalising the
        SwarmNode, SwarmRegistry, SwarmRouter, and SwarmCredit components
        and their decentralisation guarantees (\cref{sec:architecture}).
  \item We introduce the \emph{SwarmCredit Attribution} algorithm, a
        Shapley-value approximation for real-time multi-node attribution
        with trust-decay feedback and a proof-of-contribution genesis
        mechanism that resists cold-start griefing (\cref{sec:algorithm}).
  \item We analyse deployment feasibility, the bootstrap problem, security
        surface, short- and long-term benefits, and open research
        challenges, mapping a concrete path from existing HarnessAPI
        deployments to a production swarm (\cref{sec:feasibility}).
  \item We position SwarmHarness as a foundational primitive for
        autonomous distributed AI agent networks, laying a stepping stone toward
        agents that operate at internet scale with no trusted intermediary
        at any layer (\cref{sec:feasibility,sec:conclusion}).
\end{itemize}

\paragraph{Paper structure.}
\Cref{sec:related} surveys related work across agent harness frameworks,
decentralised compute, FL incentive mechanisms, and swarm intelligence.
\Cref{sec:architecture} presents the full system architecture and
decentralisation guarantees.
\Cref{sec:algorithm} formalises SwarmCredit and the cold-start solution.
\Cref{sec:feasibility} discusses feasibility, benefits, and open challenges.
\Cref{sec:conclusion} concludes.

\section{Related Work}
\label{sec:related}

\subsection{Agent Harness Frameworks}

The ReAct paradigm \citep{Yao2022react} established the pattern of
interleaving reasoning traces and tool-use actions in LLM agents, achieving
a 34\% improvement over chain-of-thought on interactive tasks.
Subsequent frameworks (LangChain, AutoGen, CrewAI) built orchestration
layers on top of this paradigm but retained a centralised coordinator that
routes tasks to agents.
AgentMesh \citep{Khanzadeh2025agentmesh} decomposes software development
into Planner, Coder, Debugger, and Reviewer agents, identifying error
propagation and context scaling as primary limitations of multi-agent
pipelines at scale.
Almasoud et al.\ \citep{Almasoud2025swarm} demonstrate that LLM agents
organised around swarm AI principles, with distinct expert personas
collaborating through dialogue, achieve 89\% alignment with human expert
scores on complex engineering problems.

Our prior work, HarnessAPI \citep{Jose2026harnessapi}, inverts the framework
dependency: a typed skill folder becomes the single source of truth from which
both HTTP endpoints and MCP tools are derived automatically.
SwarmHarness extends this single-node architecture to a decentralised
multi-node protocol, preserving full backward compatibility with existing
HarnessAPI deployments.

\subsection{Decentralised Compute Sharing}

BOINC \citep{Anderson2019boinc} is the canonical volunteer-computing platform,
demonstrating that idle consumer hardware can be harvested at scale for
scientific workloads; BOINC has collectively processed over $10^{20}$
floating-point operations across more than 5 million volunteer computers.
However, BOINC offers no monetary or credit incentive; participation is
altruistic, which limits adoption to scientific communities and is
structurally vulnerable to free-rider degradation as the network grows.
Petals \citep{Borzunov2022petals} takes a BitTorrent approach to LLM
inference, distributing BLOOM-176B across consumer GPUs.
At its peak, Petals supported over 800 active contributor nodes for interactive
inference, validating the P2P model for AI workloads, but it lacks any incentive
layer.
Atre et al.\ \citep{Atre2021volunteer} adapt asynchronous SGD to
volunteer-computing heterogeneity with 70--90\% cost reduction using preemptible
instances, showing that asynchronous execution can tolerate unreliable nodes.
Lattica \citep{Yang2025lattica} provides a decentralised AI communication substrate
using NAT traversal, CRDT-based state replication, and Distributed Hash Table
(DHT) content discovery, making it directly applicable to SwarmHarness's peer-to-peer
networking layer.
Golem \citep{Golem2018whitepaper} is the first large-scale deployed decentralised
compute marketplace, using Ethereum-based payment to incentivise compute providers;
it has processed over one million tasks via its token economy, establishing the
viability of the model at a cost of on-chain transaction overhead.

\subsection{Federated Learning Incentive Mechanisms}

FedAvg \citep{McMahan2016fedavg} established the canonical Federated Learning
(FL) protocol, enabling 10--100$\times$ communication reduction versus centralised
SGD.
However, FedAvg assumes willing participation and provides no mechanism to
recruit or reward contributors.
FedToken \citep{Pandey2022fedtoken} introduces a blockchain-backed tokenised
scheme using Shapley-value approximations to allocate tokens proportionally to
each client's model contribution during FL aggregation.
This directly informs SwarmCredit's attribution design, though we eschew the
blockchain substrate in favour of lightweight on-node ledgers.
Tian et al.\ \citep{Tian2021contract} formalise FL participation as a
contract-theoretic problem, designing two-dimensional contracts over data quality
and computation effort that satisfy individual rationality and incentive
compatibility.
Nguyen et al.\ \citep{Nguyen2025r3t} extend these contracts with time-awareness,
prioritising critical learning periods and achieving a 300\% convergence
improvement.
The comprehensive survey by Hewa Kaluannakkage and Buyya \citep{Hewa2025incentive}
taxonomises FL incentive mechanisms across economic/game-theoretic, blockchain, and
DRL-based categories, confirming that no existing mechanism simultaneously handles
skill-level granularity, real-time attribution, and blockchain-free operation.
Fan et al.\ \citep{Fan2022verfedsv} introduce VerFedSV, a communication-efficient
Shapley-value approach for vertical FL that satisfies fairness axioms while remaining
tractable; we adapt this approximation scheme for SwarmCredit.

\subsection{Multi-Agent and Swarm Intelligence}

Zhu et al.\ \citep{Zhu2022madrl} survey multi-agent deep reinforcement learning
(MARL) with communication across nine dimensions including message topology and
learning strategy.
Amato \citep{Amato2024ctde} introduces the Centralised Training, Decentralised
Execution (CTDE) paradigm, in which agents train with global information but execute
with local observations only.
SwarmHarness follows a CTDE-like philosophy: credit attribution requires global
task-outcome information, but each node's routing decisions are made locally.
Vellinger et al.\ \citep{Vellinger2025pheromones} establish the theoretical
equivalence between pheromone-mediated stigmergy in biological swarms and RL
cross-learning updates, showing that exploratory heterogeneous agents restore
collective plasticity in dynamic environments.
This equivalence motivates our use of credit and trust scores as
\emph{digital pheromones}: indirect signals that coordinate routing without
explicit communication.
Giardini et al.\ \citep{Giardini2024emergence} demonstrate that complex swarm
behaviour emerges from minimal local agent interactions, supporting our claim that
SwarmHarness can exhibit collective intelligence without explicit swarm-level
coordination protocols.
Li et al.\ \citep{Li2020ioi} propose stigmergy-based collective intelligence for
next-generation wireless networks, bridging swarm intelligence with federated
learning and showing that indirect coordination signals scale to large heterogeneous
populations.
Sajjadi et al.\ \citep{Sajjadi2024comdml} demonstrate communication-efficient
workload balancing in decentralised multi-agent learning, with faster agents
absorbing tasks from slower ones, a strategy directly applicable to SwarmRouter's load-balancing
heuristic.

\subsection{Blockchain Compute Markets}

Jaberzadeh et al.\ \citep{Jaberzadeh2023blockchain} combine IPFS, blockchain, and
smart contracts to provide tokenised incentives and on-chain penalisation of dishonest
FL contributors, demonstrating that penalty mechanisms effectively reduce free-rider
behaviour.
BrokerChain \citep{Lin2025brokerchain} extends this to a purpose-built blockchain
supporting the full AI lifecycle with an incentive-aligned economic model.
Rozemberczki et al.\ \citep{Rozemberczki2022shapley} survey Shapley-value methods
in ML across feature selection, MARL, ensemble pruning, and data valuation, providing
the canonical reference for game-theoretic contribution allocation underlying
SwarmCredit.

\subsection{Positioning}

SwarmHarness is distinguished from all prior work by the intersection of three
properties: (1) skill-first MCP nodes as the atomic unit of compute, (2) fully
decentralised DHT-based peer discovery with no trusted registry in steady state,
and (3) Shapley-fair incentive attribution without blockchain overhead.
No existing system combines all three.
The viability of each individual component is well-established at scale: Petals has
shown 800+ contributors for P2P AI inference; BOINC has demonstrated that $10^{20}$
floating-point operations can flow through volunteer hardware; Golem has settled over
one million compute tasks via a token economy.
SwarmHarness unifies these proven primitives under the skill-first MCP model,
adding the incentive layer that BOINC and Petals lack and removing the blockchain
dependency that Golem and BrokerChain require.

\section{System Architecture}
\label{sec:architecture}

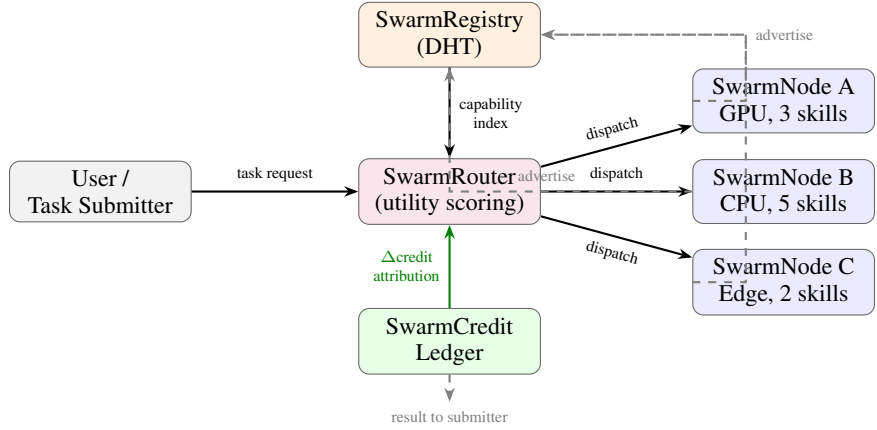
\begin{figure}[t]
\centering
\begin{tikzpicture}[
  node distance  = 1.4cm and 2.0cm,
  box/.style     = {rectangle, rounded corners=4pt, draw=black!60,
                    fill=blue!8, minimum width=2.4cm, minimum height=0.7cm,
                    font=\small, align=center},
  registry/.style= {box, fill=orange!12},
  credit/.style  = {box, fill=green!10},
  router/.style  = {box, fill=purple!10},
  user/.style    = {box, fill=gray!10},
  arr/.style     = {-{Stealth[scale=0.8]}, thick},
  darr/.style    = {arr, dashed, gray},
]

\node[user]    (user)    {User /\\Task Submitter};
\node[router, right=2.2cm of user]  (router)  {SwarmRouter\\(utility scoring)};
\node[registry, above=1.2cm of router] (registry) {SwarmRegistry\\(DHT)};
\node[box, right=2.0cm of router, yshift= 1.2cm] (n1) {SwarmNode A\\GPU, 3 skills};
\node[box, right=2.0cm of router]                (n2) {SwarmNode B\\CPU, 5 skills};
\node[box, right=2.0cm of router, yshift=-1.2cm] (n3) {SwarmNode C\\Edge, 2 skills};
\node[credit, below=1.1cm of router] (credit) {SwarmCredit\\Ledger};

\draw[arr] (user)   -- node[above, font=\tiny]{task request} (router);
\draw[arr] (router) -- node[above, font=\tiny, sloped]{dispatch} (n1);
\draw[arr] (router) -- node[above, font=\tiny]{dispatch} (n2);
\draw[arr] (router) -- node[below, font=\tiny, sloped]{dispatch} (n3);
\draw[arr] (registry) -- node[right, font=\tiny, align=center]{capability\\index} (router);
\draw[darr] (n1) -- +(-0.5,0) |- node[right, font=\tiny]{advertise} (registry);
\draw[darr] (n2) -| node[pos=0.3, above, font=\tiny]{advertise} (registry);
\draw[darr] (n3) -- +(-0.5,0) |- (registry);
\draw[arr, green!50!black] (credit) -- node[left, font=\tiny, align=center]{$\Delta$credit\\attribution} (router);
\draw[darr] (credit) -- +(0,-0.8) node[below, font=\tiny]{result to submitter};

\end{tikzpicture}
\caption{SwarmHarness end-to-end architecture. A user submits a task to the
SwarmRouter, which queries the SwarmRegistry (DHT) for candidate nodes and
scores them by a utility function. The winning node executes the skill and
returns the result. SwarmCredit attributes credit deltas to all contributing
nodes using a Shapley-value approximation; credit signals feed back into
future routing decisions as digital pheromones.}
\label{fig:architecture}
\end{figure}

SwarmHarness is a four-component protocol layered on top of HarnessAPI
\citep{Jose2026harnessapi} nodes.
\Cref{fig:architecture} shows the end-to-end system.

\subsection{SwarmNode}
\label{subsec:node}

A \emph{SwarmNode} is a HarnessAPI instance \citep{Jose2026harnessapi}
augmented with three additional responsibilities: capability advertisement,
task acceptance negotiation, and credit bookkeeping.

Formally, a SwarmNode $v$ is a tuple
$v = (\mathcal{S}_v,\, r_v,\, c_v,\, \tau_v)$,
where $\mathcal{S}_v$ is the set of skills available on node $v$, $r_v$
is the current resource vector (available VRAM, CPU fraction, network
bandwidth), $c_v$ is the SwarmCredit balance, and $\tau_v \in [0,1]$ is the
trust score accumulated over historical task completions.

On startup, a SwarmNode:
\begin{enumerate}
  \item Discovers the SwarmRegistry via bootstrap (see \cref{subsec:registry}).
  \item Publishes its capability advertisement
        $\langle \mathsf{node\_id},\, \mathcal{S}_v,\, r_v,\, \tau_v \rangle$
        to the DHT.
  \item Begins accepting inbound task requests over TLS.
\end{enumerate}

Sandboxing is enforced at the skill level: each skill handler runs in an
isolated subprocess with restricted file-system and network permissions,
consistent with the HarnessAPI execution model \citep{Jose2026harnessapi}.

\subsection{SwarmRegistry}
\label{subsec:registry}

The \emph{SwarmRegistry} is a structured overlay network (DHT) that maps
skill identifiers to lists of capable nodes.
We use a Kademlia-style DHT \citep{Maymounkov2002kademlia} where the key
is $\mathsf{SHA256}(\mathsf{skill\_name})$ and the value is a
time-to-live (TTL)-bounded list of node advertisements.

Nodes refresh their advertisement every $T_{\mathrm{refresh}}$ seconds.
A node that fails to refresh within $3 \times T_{\mathrm{refresh}}$ is
evicted from the index.
This TTL mechanism handles churn without requiring explicit leave messages.

To support capability queries beyond exact skill match (e.g., ``find nodes
with VRAM $\geq 8$\,GB''), the registry maintains a secondary in-memory
inverted index that is propagated via gossip protocol \citep{Jelasity2007gossip},
keeping it eventually consistent across the overlay.

\paragraph{Bootstrap and autodiscovery.}
A new node joining the swarm for the first time must find at least one
existing DHT participant.
SwarmHarness supports three discovery mechanisms, applied in order:
\begin{enumerate}
  \item \textbf{mDNS/local network}: nodes announce themselves over
        multicast DNS on the local subnet, requiring zero configuration for
        same-LAN deployments.
  \item \textbf{DNS seed list}: a small set of domain names
        (e.g., \texttt{seed1.swarmharness.io}) resolve to well-known
        entry-point nodes, following the same pattern used by the Bitcoin
        peer network \citep{Nakamoto2008bitcoin}.
        Multiple independent organisations operate seed nodes; no single
        operator is required, as the DNS seed list is itself replicated and
        community-governed.
  \item \textbf{Explicit peer list}: users may specify known peer addresses
        directly, enabling air-gapped or private swarm deployments.
\end{enumerate}
Once a node has joined the DHT, it learns additional peers via Kademlia's
routing table refresh and is no longer dependent on the bootstrap mechanism.
The seed nodes are \emph{entry points}, not registrars; they hold no special
authority over the overlay once a node has joined.

\subsection{SwarmRouter}
\label{subsec:router}

The \emph{SwarmRouter} is an attribution-stateless service (itself deployable
as a SwarmNode skill) that scores candidate nodes returned by the registry
and selects the best target for each task.

Given a task $T$ requiring skill $s$ and a candidate set
$\mathcal{C} = \{v_1, \ldots, v_k\}$ retrieved from the DHT, the router
scores each candidate with a utility function:

\begin{equation}
  U(v, T) = w_1 \cdot \mathbf{1}[s \in \mathcal{S}_v]
           + w_2 \cdot (1 - \ell_v)
           + w_3 \cdot (1 - d_v / d_{\max})
           + w_4 \cdot \tau_v
  \label{eq:utility}
\end{equation}

where $\ell_v \in [0,1]$ is the current load fraction of node $v$,
$d_v$ is the round-trip latency to $v$, $d_{\max}$ is a configurable
latency ceiling, and $\tau_v$ is the trust score.
Weights $w_1, \ldots, w_4$ sum to one and are tunable per-deployment.

The router dispatches to $\arg\max_v U(v,T)$ with a random tie-breaking
fallback to the next-best candidate on failure.
For tasks that benefit from redundancy (e.g., ensemble inference), the
router may dispatch to the top-$K$ nodes and merge results.

Multiple SwarmRouter instances may operate concurrently; a submitter
selects any router from the DHT index, since routers are themselves listed as
a skill type.
Because credit attribution is \emph{signed by the task submitter}, not by
the router, a malicious or crashed router cannot corrupt the credit ledger.
Router diversity is thus a safety property: the system degrades gracefully
if individual routers fail, and submitters can freely switch routers without
losing credit history.

NAT traversal is handled by Lattica-style hole-punching
\citep{Yang2025lattica}; nodes behind symmetric NATs fall back to a TURN
relay hosted by a volunteered SwarmNode with public connectivity.

\subsection{SwarmCredit Ledger}
\label{subsec:ledger}

SwarmCredit is a lightweight, blockchain-free credit system.
Each node maintains a local ledger of its own credit balance $c_v$.
Credit deltas $\Delta c_v$ are computed after task completion using the
SwarmCredit Attribution algorithm (\cref{sec:algorithm}) and communicated
directly to participating nodes over the same TLS channel used for task
delivery.

Credit is not a cryptocurrency; it is not globally consistent, and nodes
are not expected to trade it externally.
Its purpose is to modulate routing priority: the utility function
(\cref{eq:utility}) incorporates trust score $\tau_v$, which accumulates as
a function of historical credit attributions.
Nodes with sustained positive attributions earn higher $\tau_v$ and attract
more tasks, while idle or low-quality nodes lose routing priority.

\subsection{Decentralisation Guarantees}
\label{subsec:decentralisation}

We distinguish two levels of decentralisation:

\textbf{Operational decentralisation} means no single actor can unilaterally
control task routing or credit attribution.
SwarmHarness achieves this from Phase 1 (the initial deployment): the utility
function (\cref{eq:utility}) runs locally at each router instance; credit
attribution is signed by the task submitter; the DHT has no privileged
participants once a node has joined.

\textbf{Administrative decentralisation} means no single actor controls
registry membership or the bootstrap process.
SwarmHarness achieves this progressively:
\begin{itemize}
  \item \emph{Phase 1 (alpha)}: a single reference seed server operated by
        the project team; acceptable for early adopters who trust the project.
  \item \emph{Phase 2 (federated)}: multiple independent seed operators publish
        DNS records; the seed list is community-governed and openly auditable, so
        no single operator can deny network access.
  \item \emph{Phase 3 (fully decentralised)}: bootstrap via mDNS (local),
        trusted-peer exchange (social), and direct peer lists (technical);
        DNS seeds become optional convenience, not a requirement.
        At Phase 3, SwarmHarness has no administratively central component.
\end{itemize}

This roadmap is analogous to Bitcoin's transition from a single seed IRC channel
to a distributed DNS-seed network to a fully peer-exchanged overlay
\citep{Nakamoto2008bitcoin}.
The key property is that \emph{operational decentralisation is present from day
one}; administrative decentralisation is a phased improvement, not a prerequisite
for correct operation.

\section{SwarmCredit Attribution Algorithm}
\label{sec:algorithm}

\subsection{Problem Formulation}

Let a task $T$ be executed by a coalition of nodes
$\mathcal{N} = \{N_1, N_2, \ldots, N_k\}$.
For multi-node tasks (e.g., ensemble inference, pipeline chains), each node
contributes a partial service.
Define the characteristic function $v: 2^{\mathcal{N}} \to \mathbb{R}$
where $v(S)$ is the quality of task outcome achievable by coalition $S$
alone, measured against a quality signal $q \in [0,1]$ derived from the
task response (e.g., user rating, downstream model score, or
latency-SLA compliance).
For single-node tasks, $k = 1$ and credit attribution reduces to direct
assignment.

We seek a credit allocation
$\boldsymbol{\Delta c} = (\Delta{}c_1, \ldots, \Delta{}c_k)$
satisfying:
\begin{itemize}
  \item \textbf{Efficiency:} $\sum_i \Delta{}c_i = C(T)$,
        the total credit pool for task $T$.
  \item \textbf{Fairness:} allocation follows the Shapley value
        \citep{Rozemberczki2022shapley}.
  \item \textbf{Individual rationality:} $\Delta{}c_i \geq 0$ for all $i$
        when the coalition produces positive quality.
\end{itemize}

The difficulty lies in computing fair attributions when contributions are
order-dependent.
A na\"{i}ve proportional split (dividing $C(T)$ equally among all $k$ nodes)
ignores the fact that one node may be responsible for most of the quality
gain while another contributes only marginally: a node that handles the
bottleneck step in a pipeline chain should receive more credit than one that
merely reformats its output.
The Shapley value resolves this by averaging each node's \emph{marginal
contribution} over all possible orderings in which it could join the
coalition, making the attribution the unique allocation satisfying the
four axioms of symmetry, efficiency, null-player, and additivity
\citep{Rozemberczki2022shapley}.
Exact computation requires evaluating $v(S)$ for every subset $S \subseteq
\mathcal{N}$, giving $O(2^k)$ evaluations; this is intractable for large
coalitions but manageable for the small $k$ values that arise in practice
(at most $k = 10$ for ensemble inference), and can be approximated
efficiently via random permutation sampling.

\subsection{SwarmCredit Attribution}

Algorithm~\ref{alg:swarmcredit} implements Shapley-value credit attribution
via Monte Carlo permutation sampling \citep{Rozemberczki2022shapley}, resolving
the $O(2^k)$ intractability with $M$ random permutation draws.
The algorithm proceeds in four logical steps: (1)~estimate each node's Shapley
value $\phi_i$ by averaging marginal contributions across $M$ random orderings;
(2)~normalise the positive Shapley values to the credit pool $C(T)$, enforcing
the Efficiency axiom exactly; (3)~update each node's trust score $\tau_i$ based
on whether it received a positive credit attribution; and (4)~deduct the task
cost from the submitter's balance.
The submitter deduction occurs last so that concurrent tasks drawing from the
same credit pool are each individually validated before any balance is modified,
preventing double-spending in overlapping task windows.

\begin{algorithm}[t]
\SetAlgoLined
\KwIn{Task $T$; contributing nodes $\mathcal{N} = \{N_1,\ldots,N_k\}$;
       quality signal $q \in [0,1]$; credit pool $C(T) > 0$;
       trust learning rate $\alpha \in (0,1)$; sample count $M$}
\KwOut{Credit deltas $\Delta{}c_1,\ldots,\Delta{}c_k$;
        updated trust scores $\tau_1,\ldots,\tau_k$}

\BlankLine
\tcp{Step 1 — estimate Shapley values via random permutation sampling}
$\phi_i \leftarrow 0$ for all $i$\;
\For{$m = 1$ \KwTo $M$}{
  $\pi \leftarrow \mathrm{RandomPermutation}(\mathcal{N})$\;
  \For{$i = 1$ \KwTo $k$}{
    $S_{\text{pre}} \leftarrow \{N_j : \pi^{-1}(j) < \pi^{-1}(i)\}$\;
    $\phi_i \leftarrow \phi_i + q(S_{\text{pre}} \cup \{N_i\}) - q(S_{\text{pre}})$\;
  }
}
$\phi_i \leftarrow \phi_i / M$ for all $i$\;

\BlankLine
\tcp{Step 2 — normalise to credit pool}
$\Phi \leftarrow \sum_{i=1}^{k} \max(\phi_i, 0)$\;
\eIf{$\Phi > 0$}{
  $\Delta{}c_i \leftarrow \dfrac{\max(\phi_i, 0)}{\Phi} \cdot C(T)$\;
}{
  $\Delta{}c_i \leftarrow C(T) / k$ (uniform fallback)\;
}

\BlankLine
\tcp{Step 3 — update trust scores}
\For{$i = 1$ \KwTo $k$}{
  \eIf{$\Delta{}c_i > 0$}{
    $\tau_i \leftarrow \tau_i + \alpha \cdot (1 - \tau_i)$\;
  }{
    $\tau_i \leftarrow \tau_i - \alpha \cdot \tau_i$\;
  }
}

\BlankLine
\tcp{Step 4 — deduct task cost from submitter}
$c_{\text{submitter}} \leftarrow c_{\text{submitter}} - C(T)$\;

\KwRet{$(\Delta{}c_1,\ldots,\Delta{}c_k),\; (\tau_1,\ldots,\tau_k)$}
\caption{SwarmCredit Attribution}
\label{alg:swarmcredit}
\end{algorithm}

Step~2 enforces Efficiency exactly: the normalisation ensures $\sum_i \Delta{}c_i = C(T)$
regardless of the sampled Shapley estimates.
Individual Rationality is guaranteed structurally: the $\max(\phi_i, 0)$ clamping
in Step~2 ensures no node receives a negative credit delta, and the uniform fallback
applies only when every sampled Shapley value is non-positive (indicating a task where
no node improved quality above the empty-coalition baseline).
Fairness is approximated with standard error $O(1/\sqrt{M})$, converging to the true
Shapley value as $M$ increases.

\subsection{Quality Signal}

The quality signal $q(S)$ must be computable without executing each
sub-coalition independently, since that would be computationally
prohibitive.
We approximate $q(S)$ using leave-one-out proxies: in a pipeline
chain, removing $N_i$ degrades throughput proportionally to its processing
time share; in ensemble inference, removing $N_i$ increases variance
proportionally to its historical accuracy differential.
This approximation is similar in spirit to VerFedSV \citep{Fan2022verfedsv},
which computes communication-efficient Shapley values for vertical FL by
treating vertical partitions as leave-one-out coalitions.

\subsection{Credit Drain for Idle Nodes}

Nodes that do not accept tasks do not lose credit instantly; passive
holding is permitted.
However, the utility function (\cref{eq:utility}) incorporates $\tau_v$,
and trust scores decay toward zero when a node is not selected for routing.
The decay rule is:

\begin{equation}
  \tau_v \leftarrow \tau_v \cdot (1 - \beta)^{\Delta t / T_0}
  \label{eq:trustdecay}
\end{equation}

where $\beta$ is the per-period decay rate, $\Delta t$ is the time since
the node's last accepted task, and $T_0$ is the normalisation period (e.g.,
24 hours).
A node that never accepts tasks sees $\tau_v \to 0$, reducing $U(v,T)$
toward the capability-only term and eventually starving it of routing
traffic entirely.
Node operators are thus incentivised to keep their nodes responsive.

\subsection{Cold-Start and Genesis Credits}
\label{subsec:coldstart}

A new SwarmNode joining the network faces a cold-start problem: it has zero
credits and zero trust, making it unlikely to be selected by the router
despite having available capacity.
Similarly, a new task submitter has zero credits and cannot submit their
first task.
We address both cases with a \emph{proof-of-contribution genesis} mechanism.

A node receives a genesis credit endowment $c_0$ sufficient to submit
$k_0$ tasks.
However, the genesis pool is \emph{locked} until the node has successfully
served at least one task that was countersigned by the task submitter, which
constitutes proof of contribution.
The countersignature requirement prevents \emph{genesis credit griefing},
in which a malicious actor repeatedly creates new node identities
(each with a fresh $c_0$ endowment) to harvest credits without contributing.
A Sybil attacker must actually execute at least one legitimate task per
identity before accessing any genesis funds, which imposes a cost proportional
to the volume of free-riding attempted.

For new task submitters (with no served tasks), a \emph{genesis task grant}
allows one free task submission.
After this first task, the submitter must earn credits by running a node or
purchase them from other nodes via out-of-band bilateral agreements.
The genesis grant is rate-limited per IP prefix and per signing keypair,
making sustained abuse expensive without genuine participation.

Together, these mechanisms ensure that the network bootstraps from zero
without creating exploitable free-rider subsidies: every credit eventually
traces back to a verified compute contribution.

\subsection{Complexity and Approximation Bounds}

Exact Shapley computation requires $O(2^k)$ coalition evaluations, which is
intractable for $k > 20$.
The random permutation sampling approximation (\cref{alg:swarmcredit},
Step~1) converges to the true Shapley value with standard error
$O(1/\sqrt{M})$ \citep{Rozemberczki2022shapley}.
In practice, SwarmHarness tasks involve $k \leq 5$ nodes for pipelined
skills and $k \leq 10$ for ensemble inference, making $M = 100$ samples
sufficient for sub-1\% standard error.
For the common single-node case ($k=1$), attribution is exact with $O(1)$
cost.

\subsection{Swarm Emergence via Credit Pheromones}

The credit and trust signals in SwarmHarness function as digital pheromones
in the sense of Vellinger et al.\ \citep{Vellinger2025pheromones}: indirect
signals that modify the effective environment for other agents without direct
communication.
Concretely:

\begin{itemize}
  \item High-credit, high-trust nodes attract more routing traffic, generating
        more credit and further increasing routing priority.
        This positive feedback loop concentrates compute on reliable, high-skill
        nodes.
  \item When a popular node approaches full load, its load fraction $\ell_v$
        penalises its utility score (\cref{eq:utility}), redirecting traffic
        to less-loaded nodes.
        This negative feedback loop prevents monopolisation and maintains
        diversity in the swarm.
  \item Over time, nodes specialise: a node receiving many inference tasks
        optimises its skill set toward inference, increasing its quality signal
        and credit per task.
        Nodes that perform poorly on a skill type gradually lose traffic on
        that skill, analogous to pheromone evaporation on low-quality paths
        in ant colony optimisation \citep{Li2020ioi}.
\end{itemize}

The combination of positive and negative feedback produces emergent load
distribution and skill specialisation without any explicit coordination
protocol, constituting swarm intelligence at the network level.

\section{Feasibility and Open Challenges}
\label{sec:feasibility}

\subsection{Deployment Path}

SwarmHarness is designed for incremental adoption.
Phase~1 begins with existing HarnessAPI deployments: node operators
install the swarm extension package (a pip-installable addon exposing
the DHT join and credit ledger APIs) and opt in to the registry.
No changes to existing skill folders are required.
Phase~2 introduces a federated DNS seed network in which multiple independent
organisations publish seed records, following the Bitcoin DNS-seed model.
Phase~3 aims for full bootstrap independence via mDNS, trusted-peer
exchange, and direct peer lists, with DNS seeds becoming optional
convenience rather than a requirement (see \cref{subsec:decentralisation}).

\subsection{Short- and Long-Term Benefits}
\label{subsec:benefits}

\begin{description}
  \item[\textbf{Short-term (Phase~1, months 0--6).}]
    Individual developers running existing HarnessAPI nodes can immediately
    enrol in the swarm, earning credits for skills that other community members
    need.
    A developer with a GPU running a local inference skill earns credits that
    can be spent on summarisation or code-analysis skills running on a
    colleague's server, forming a barter economy that reduces cloud spend
    without any infrastructure changes.
    Small research teams can share compute across lab machines, dispatching
    ML evaluation jobs to whichever machine is currently idle, without managing
    a job queue or a shared cluster account.
    The barrier to entry is a single \texttt{pip install} and one configuration
    command.

  \item[\textbf{Medium-term (Phase~2, months 6--24).}]
    As the swarm grows, the credit economy self-regulates supply and demand for
    popular skills.
    Research institutions can seed community swarms by operating well-provisioned
    nodes, providing a public good analogous to open arXiv preprint servers.
    Compute-intensive workloads such as hyperparameter sweeps, large inference
    batches, and dataset preprocessing can be dispatched across dozens of
    volunteer nodes without renting cloud instances.
    The credit system ensures that high-demand skills attract more contributors,
    while low-demand skills are not artificially over-provisioned.

  \item[\textbf{Long-term (Phase~3, years 2+).}]
    At full administrative decentralisation, SwarmHarness becomes a
    general-purpose peer-to-peer compute mesh with no privileged actors.
    Any AI workload, including inference, fine-tuning, data processing, and
    agentic pipelines, can be dispatched across the swarm without cloud
    intermediaries.
    Most compellingly, SwarmHarness provides the substrate for \emph{autonomous
    distributed AI agents}: software agents that decompose complex goals into
    subtasks, dispatch each subtask to the best available SwarmNode, settle
    credits automatically, and aggregate results, all without human
    intermediation.
    Such agents would treat SwarmHarness the way HTTP clients treat the internet:
    as a reliable, anonymous, incentive-aligned transport layer for computation.
\end{description}

\subsection{The Bootstrap Problem}

Decentralised compute networks face a chicken-and-egg problem: new users
will not submit tasks to an empty swarm, and node operators will not join
a swarm with no task demand.
We address this with three complementary mechanisms.
First, the \emph{proof-of-contribution genesis} scheme described in
\cref{subsec:coldstart} ensures new nodes can access the network immediately
while preventing free-rider exploitation of genesis credits.
Second, \emph{task subsidies from anchor operators}: research institutions
and developers who benefit from the swarm are encouraged to run
well-provisioned nodes that contribute excess capacity, seeding demand,
mirroring the strategy used by Golem \citep{Golem2018whitepaper} in its
early deployment.
Third, \emph{local swarms}: in Phase~1, small private swarms (e.g., a single
research group's machines) can operate entirely without public seed nodes,
using mDNS for autodiscovery, with credits transferable to the public swarm
as Phase~2 seeds come online.

\subsection{Security and Privacy}

\textbf{Node authentication.}
All inter-node communication uses mutual TLS with certificate pinning.
Node identities are derived from Ed25519 keypairs generated at first run, and
the DHT stores each node's public key alongside its capability advertisement,
enabling downstream routers and submitters to verify task responses and credit
attribution signatures.

\textbf{Skill sandboxing.}
Each skill handler executes in a subprocess with Linux namespaces (or macOS
sandbox profiles) restricting file-system access to the skill directory and
network access to whitelisted endpoints, following the sandboxed execution
model in HarnessAPI \citep{Jose2026harnessapi}.

\textbf{Sybil resistance.}
Sybil attacks, where an adversary creates many low-cost identities to
accumulate credit or skew routing, are a known challenge in decentralised
systems \citep{Jaberzadeh2023blockchain}.
SwarmHarness mitigates this with proof-of-work during node registration
(a lightweight SHA-256 puzzle, not a full consensus protocol) combined with
the trust score's historical dependence: newly registered nodes start with
$\tau_v = 0$ and must earn trust through completed tasks, so identity cycling
imposes a real compute cost on attackers.

\textbf{Collusion and credit inflation.}
A set of colluding nodes could fabricate task completions to inflate their
credit balances.
We mitigate this by requiring the task submitter (who holds the quality
signal $q$) to countersign the credit attribution before it is applied.
A submitter who does not countersign within a timeout receives no future
routing priority, creating a mutual deterrent; systematic collusion therefore
requires compromising both colluding nodes and their submitters simultaneously,
significantly raising the attack cost.

\textbf{Privacy.}
Task payloads are transmitted directly between the submitter and the executing
node over TLS and are never persisted by the router or registry.
The SwarmCredit ledger stores only numeric credit deltas and timestamps, not
task content, and node capability advertisements contain only skill names and
resource metrics, not identifying information beyond the node's public key.

\subsection{Open Research Challenges}

\textbf{Credit valuation.}
The current SwarmCredit scheme is ordinal: it determines relative routing
priority, not an exchange rate.
An open question is whether credits should be exchangeable for external
resources (cloud compute, API tokens) and, if so, how to prevent credit
inflation; market-clearing prices that reflect supply and demand for each
skill type would improve efficiency but require a distributed auction
mechanism \citep{Tian2021contract}.

\textbf{Heterogeneous quality signals.}
Quality signal design is task-specific.
For subjective tasks (creative writing, open-ended question answering),
automated quality proxies are unreliable, and human-in-the-loop rating
introduces latency and cost.
Extending SwarmCredit to handle uncertain or delayed quality signals is an
important open problem.

\textbf{Byzantine fault tolerance.}
The current protocol assumes rational (self-interested but non-malicious)
nodes.
Extending to a Byzantine fault-tolerant model, where some nodes may return
incorrect results intentionally, would require cryptographic verification
schemes such as zero-knowledge proofs of correct execution, at a significant
performance cost.

\textbf{Dynamic skill pricing.}
All tasks in the current model draw from a credit pool $C(T)$ set by the
submitter.
Allowing nodes to advertise per-skill prices and enabling market-clearing
would improve resource allocation efficiency, but doing so in a fully
decentralised setting requires an auction protocol that remains an active
area of research in mechanism design \citep{Tian2021contract}.

\subsection{Future Directions}

\textbf{Skill marketplaces.}
Nodes may publish not just skill availability but credit-per-call pricing,
enabling a skill marketplace in which submitters compare offerings before
dispatching.
Differential pricing would allow scarce high-performance skills (e.g., large
GPU inference) to attract more contributors, while common low-cost skills
(e.g., text classification) remain competitively priced.

\textbf{Federated fine-tuning.}
Combining SwarmHarness with FL incentive mechanisms
\citep{Pandey2022fedtoken,Hewa2025incentive} would allow nodes to contribute
not only inference compute but also local training data, enabling
privacy-preserving federated fine-tuning of foundation models across the
swarm without any central parameter server.

\textbf{Towards autonomous distributed agents.}
The long-term vision for SwarmHarness extends beyond human-initiated task
dispatch.
As autonomous AI agents (programs that independently pursue goals by composing
skills) become more capable, they will require a compute substrate that is
always available, fairly priced, and resistant to single points of failure.
SwarmHarness provides exactly this: a node that an agent can query for
available skills, dispatch subtasks to, and settle credits with, all through
the same MCP interface that human-facing tools already use.
An agent operating within SwarmHarness is indistinguishable from a human
submitter; it creates a keypair, earns credits by running a node, and spends
them on compute it needs, with no special runtime, privileged API, or trusted
intermediary required.
This makes SwarmHarness a foundational primitive for the next generation of
distributed autonomous agent systems: one where the network itself provides
the incentive alignment that today's centralised agent platforms impose through
terms of service.

\section{Conclusion}
\label{sec:conclusion}

We presented \textbf{SwarmHarness}, a decentralised incentive-aligned protocol
for organising HarnessAPI skill nodes into a self-governing compute swarm.
The system addresses a gap that existing approaches leave open: decentralised
compute sharing with a fair, real-time incentive mechanism that requires no
blockchain, no central coordinator, and no changes to existing skill
implementations.

SwarmHarness contributes three technical elements.
The \emph{SwarmRegistry} provides Kademlia DHT-based peer discovery with
three complementary bootstrap mechanisms (mDNS, DNS seeds, and explicit
peer lists), handling churn through TTL-bounded advertisements and eventual
consistency via gossip propagation.
The \emph{SwarmRouter} dispatches tasks using a utility function that jointly
optimises capability match, load, latency, and trust, with multiple concurrent
router instances providing fault tolerance through diversity; submitter-signed
credit attribution ensures that no router can corrupt the credit ledger.
\emph{SwarmCredit Attribution} (\cref{alg:swarmcredit}) allocates compute
credits to contributing nodes using a Shapley-value approximation with
$O(1/\sqrt{M})$ convergence, while a proof-of-contribution genesis mechanism
eliminates cold-start free-riding.
A trust-decay rule (\cref{eq:trustdecay}) continuously penalises idle and
unreliable nodes, keeping the routing graph adaptive.

The combination of positive and negative feedback from the credit and trust
signals produces emergent swarm behaviour (specialisation, load balancing,
and collective intelligence) without any explicit coordination protocol.
This connects SwarmHarness to the theoretical framework of digital pheromones
\citep{Vellinger2025pheromones} and validates the claim that the network
exhibits genuine swarm intelligence at scale.

Beyond individual compute sharing, SwarmHarness lays the groundwork for
autonomous AI agents that operate entirely within the swarm, hiring compute,
routing subtasks, and settling credits without human intermediation, a step
toward self-sustaining distributed intelligence at internet scale.
An agent interacting with SwarmHarness is indistinguishable from a human
submitter: it uses the same MCP interface, the same credit economy, and the
same trust model.
No privileged runtime is needed.

A reference implementation of SwarmHarness as a HarnessAPI extension is
planned as open-source software.
We invite the research community to join the genesis swarm, contribute
compute, and extend the protocol toward skill marketplaces, federated
fine-tuning, and fully autonomous distributed agent networks.

\bibliographystyle{unsrt}
\bibliography{references}

@misc{Yao2022react,
  author       = {Shunyu Yao and Jeffrey Zhao and Dian Yu and Nan Du and
                  Izhak Shafran and Karthik Narasimhan and Yuan Cao},
  title        = {{ReAct}: Synergizing Reasoning and Acting in Language Models},
  year         = {2022},
  howpublished = {arXiv:\texttt{2210.03629}},
  note         = {arXiv:2210.03629},
}

@misc{Jose2026harnessapi,
  author       = {Edwin Jose},
  title        = {{HarnessAPI}: A Skill-First Framework for Unified
                  Streaming {APIs} and {MCP} Tools},
  year         = {2026},
  howpublished = {arXiv preprint},
  note         = {arXiv preprint, May 2026},
}

@misc{Khanzadeh2025agentmesh,
  author       = {Milad Khanzadeh},
  title        = {{AgentMesh}: A Cooperative Multi-Agent Generative {AI}
                  Framework for Software Development Automation},
  year         = {2025},
  howpublished = {arXiv:\texttt{2507.19902}},
  note         = {arXiv:2507.19902},
}

@misc{Almasoud2025swarm,
  author       = {Mona Almasoud and others},
  title        = {Harnessing Multi-Agent {LLMs} for Complex Engineering
                  Problem-Solving: A Swarm Intelligence Approach},
  year         = {2025},
  howpublished = {arXiv:\texttt{2501.01205}},
  note         = {arXiv:2501.01205; IEEE EDUCON 2025},
}

@misc{Borzunov2022petals,
  author       = {Alexander Borzunov and Dmitry Baranchuk and Tim Dettmers and
                  Max Ryabinin and Younes Belkada and Artem Chumachenko and
                  Pavel Samygin and Colin Raffel},
  title        = {Petals: Collaborative Inference and Fine-tuning of Large Models},
  year         = {2022},
  howpublished = {arXiv:\texttt{2209.01188}},
  note         = {arXiv:2209.01188},
}

@misc{Anderson2019boinc,
  author       = {David P. Anderson},
  title        = {{BOINC}: A Platform for Volunteer Computing},
  year         = {2019},
  howpublished = {arXiv:\texttt{1903.01699}},
  note         = {arXiv:1903.01699},
}

@misc{Yang2025lattica,
  author       = {Yang and others},
  title        = {Lattica: A Decentralized Cross-{NAT} Communication Framework
                  for Scalable {AI} Inference and Training},
  year         = {2025},
  howpublished = {arXiv:\texttt{2510.00183}},
  note         = {arXiv:2510.00183},
}

@misc{Atre2021volunteer,
  author       = {Madhurima Atre and Bhavya Jha and Yogesh Simmhan},
  title        = {Distributed Deep Learning using Volunteer Computing-Like Paradigm},
  year         = {2021},
  howpublished = {arXiv:\texttt{2103.08894}},
  note         = {arXiv:2103.08894; IEEE IPDPSW 2021},
}

@inproceedings{Lin2025brokerchain,
  author    = {Jianru Lin},
  title     = {Decentralized {AI} Infrastructure Built on {BrokerChain}},
  booktitle = {Proceedings of the ACM},
  year      = {2025},
  doi       = {10.1145/3769698.3771235},
}

@misc{Golem2018whitepaper,
  author       = {{Golem Factory}},
  title        = {Golem: A Decentralised Supercomputer},
  year         = {2018},
  howpublished = {arXiv:\texttt{1801.04024}},
  note         = {arXiv:1801.04024; Technical whitepaper},
}

@misc{McMahan2016fedavg,
  author       = {H. Brendan McMahan and Eider Moore and Daniel Ramage and
                  Seth Hampson and Blaise {Ag{\"u}era y Arcas}},
  title        = {Communication-Efficient Learning of Deep Networks from
                  Decentralized Data},
  year         = {2023},
  howpublished = {arXiv:\texttt{1602.05629}},
  note         = {arXiv:1602.05629},
}

@misc{Pandey2022fedtoken,
  author       = {Shashi Raj Pandey and Lam Duc Nguyen and Petar Popovski},
  title        = {{FedToken}: Tokenized Incentives for Data Contribution
                  in Federated Learning},
  year         = {2022},
  howpublished = {arXiv:\texttt{2209.09775}},
  note         = {arXiv:2209.09775},
}

@misc{Tian2021contract,
  author       = {Mengmeng Tian and Yuxin Chen and Yuan Liu and Zehui Xiong
                  and Cyril Leung and Chunyan Miao},
  title        = {A Contract Theory Based Incentive Mechanism for
                  Federated Learning},
  year         = {2021},
  howpublished = {arXiv:\texttt{2108.05568}},
  note         = {arXiv:2108.05568},
}

@misc{Nguyen2025r3t,
  author       = {Nguyen and others},
  title        = {Right Reward Right Time for Federated Learning},
  year         = {2025},
  howpublished = {arXiv:\texttt{2503.07869}},
  note         = {arXiv:2503.07869},
}

@misc{Hewa2025incentive,
  author       = {Chanuka A. S. Hewa Kaluannakkage and Rajkumar Buyya},
  title        = {Incentive-Based Federated Learning: Architectural Elements
                  and Future Directions},
  year         = {2025},
  howpublished = {arXiv:\texttt{2510.14208}},
  note         = {arXiv:2510.14208},
}

@misc{Fan2022verfedsv,
  author       = {Zhenan Fan and Huang Fang and Xinglu Wang and Zirui Zhou
                  and Jian Pei and Michael P. Friedlander and Yong Zhang},
  title        = {Fair and Efficient Contribution Valuation for Vertical
                  Federated Learning},
  year         = {2022},
  howpublished = {arXiv:\texttt{2201.02658}},
  note         = {arXiv:2201.02658},
}

@misc{Zhu2022madrl,
  author       = {Changxi Zhu and Mehdi Dastani and Shihan Wang},
  title        = {A Survey of Multi-Agent Deep Reinforcement Learning
                  with Communication},
  year         = {2022},
  howpublished = {arXiv:\texttt{2203.08975}},
  note         = {arXiv:2203.08975},
}

@misc{Amato2024ctde,
  author       = {Christopher Amato},
  title        = {An Introduction to Centralized Training for Decentralized
                  Execution in Cooperative Multi-Agent Reinforcement Learning},
  year         = {2024},
  howpublished = {arXiv:\texttt{2409.03052}},
  note         = {arXiv:2409.03052},
}

@misc{Vellinger2025pheromones,
  author       = {Vellinger and Antonic and Tuci},
  title        = {From Pheromones to Policies: Reinforcement Learning for
                  Engineered Biological Swarms},
  year         = {2025},
  howpublished = {arXiv:\texttt{2509.20095}},
  note         = {arXiv:2509.20095},
}

@misc{Giardini2024emergence,
  author       = {Giardini and Hardy and {da Cunha}},
  title        = {Evolving Neural Networks Reveal Emergent Collective Behavior
                  from Minimal Agent Interactions},
  year         = {2024},
  howpublished = {arXiv:\texttt{2410.19718}},
  note         = {arXiv:2410.19718},
}

@misc{Li2020ioi,
  author       = {Li and others},
  title        = {Internet of Intelligence: Collective Advantage for
                  Communications},
  year         = {2020},
  howpublished = {arXiv:\texttt{1905.00719}},
  note         = {arXiv:1905.00719},
}

@misc{Sajjadi2024comdml,
  author       = {Seyed Mahmoud Sajjadi Mohammadabadi and Lei Yang and
                  Feng Yan and Junshan Zhang},
  title        = {Communication-Efficient Training Workload Balancing for
                  Decentralized Multi-Agent Learning},
  year         = {2024},
  howpublished = {arXiv:\texttt{2405.00839}},
  note         = {arXiv:2405.00839; IEEE ICDCS 2024},
  doi          = {10.1109/ICDCS60910.2024.00069},
}

@misc{Jaberzadeh2023blockchain,
  author       = {Jaberzadeh and others},
  title        = {Blockchain-Based Federated Learning: Incentivizing Data
                  Sharing and Penalizing Dishonest Behavior},
  year         = {2023},
  howpublished = {arXiv:\texttt{2307.10492}},
  note         = {arXiv:2307.10492},
}

@misc{Rozemberczki2022shapley,
  author       = {Benedek Rozemberczki and Lauren Watson and P{\'e}ter Bayer and
                  Hao-Tsung Yang and Oliv{\'e}r Kiss and Sebastian Nilsson and
                  Rik Sarkar},
  title        = {The Shapley Value in Machine Learning},
  year         = {2022},
  howpublished = {arXiv:\texttt{2202.05594}},
  note         = {arXiv:2202.05594},
}

@misc{Nakamoto2008bitcoin,
  author       = {Satoshi Nakamoto},
  title        = {Bitcoin: A Peer-to-Peer Electronic Cash System},
  year         = {2008},
  howpublished = {\url{https://bitcoin.org/bitcoin.pdf}},
  note         = {Accessed: May 2026},
}

@inproceedings{Maymounkov2002kademlia,
  author    = {Petar Maymounkov and David Mazi{\`e}res},
  title     = {Kademlia: A Peer-to-Peer Information System Based on the
               {XOR} Metric},
  booktitle = {Revised Papers from the 1st International Workshop on
               Peer-to-Peer Systems ({IPTPS})},
  year      = {2002},
  pages     = {53--65},
  publisher = {Springer},
  doi       = {10.1007/3-540-45748-8_5},
}

@article{Jelasity2007gossip,
  author  = {M{\'a}rk Jelasity and Alberto Montresor and Ozalp Babaoglu},
  title   = {Gossip-Based Aggregation in Large Dynamic Networks},
  journal = {ACM Transactions on Computer Systems},
  year    = {2005},
  volume  = {23},
  number  = {3},
  pages   = {219--252},
  doi     = {10.1145/1082469.1082470},
}

\end{document}